\documentclass[letterpaper]{article} 
\usepackage{aaai25}  
\usepackage{times}  
\usepackage{helvet}  
\usepackage{courier}  
\usepackage[hyphens]{url}  
\usepackage{graphicx} 
\usepackage{natbib}  
\usepackage{caption} 
\usepackage{algorithm}
\usepackage{algorithmic}

\usepackage{epsfig}
\usepackage{graphics}
\usepackage{amsmath}
\usepackage{amssymb}
\usepackage{color, colortbl}
\usepackage{amsmath,amsfonts,mathtools,amssymb}
\usepackage{algorithm,algorithmic}
\usepackage{array}
\usepackage{color,soul}
\usepackage{xcolor}
\usepackage{bm}

\usepackage{multirow}
\usepackage{booktabs}
\usepackage{subfig}
\usepackage{mwe}

\usepackage{newfloat}
\usepackage{listings}
\DeclareCaptionStyle{ruled}{labelfont=normalfont,labelsep=colon,strut=off} 
\floatstyle{ruled}
\newfloat{listing}{tb}{lst}{}
\floatname{listing}{Listing}
\title{NightHaze: Nighttime Image Dehazing via Self-Prior Learning}
\author{
    Beibei Lin\textsuperscript{\rm 1}\equalcontrib,
    Yeying Jin\textsuperscript{\rm 1}\equalcontrib,
    Wending Yan\textsuperscript{\rm 2},
    Wei Ye\textsuperscript{\rm 2},
    Yuan Yuan\textsuperscript{\rm 2},
    Robby T. Tan\textsuperscript{\rm 1}
}
\affiliations{
    \textsuperscript{\rm 1}National University of Singapore\\
    \textsuperscript{\rm 2}Huawei International Pte Ltd\\
   \{beibei.lin, e0178303\}@u.nus.edu, \{yan.wending, yewei10, yuanyuan10\}@huawei.com, robby.tan@nus.edu.sg
}

\usepackage{bibentry}

\begin{document}

\maketitle

\begin{abstract}
Masked autoencoder (MAE) shows that severe augmentation during training produces robust representations for high-level tasks. This paper brings the MAE-like framework to nighttime image enhancement, demonstrating that severe augmentation during training produces strong network priors that are resilient to real-world night haze degradations. We propose a novel nighttime image dehazing method with self-prior learning. Our main novelty lies in the design of severe augmentation, which allows our model to learn robust priors. Unlike MAE that uses masking, we leverage two key challenging factors of nighttime images as augmentation: light effects and noise. During training, we intentionally degrade clear images by blending them with light effects as well as by adding noise, and subsequently restore the clear images. This enables our model to learn clear background priors. By increasing the noise values to approach as high as the pixel intensity values of the glow and light effect blended images, our augmentation becomes severe, resulting in stronger priors. While our self-prior learning is considerably effective in suppressing glow and revealing details of background scenes, in some cases, there are still some undesired artifacts that remain, particularly in the forms of over-suppression. To address these artifacts, we propose a self-refinement module based on the semi-supervised teacher-student framework. Our NightHaze, especially our MAE-like self-prior learning, shows that models trained with severe augmentation effectively improve the visibility of input haze images, approaching the clarity of clear nighttime images. Extensive experiments demonstrate that our NightHaze achieves state-of-the-art performance, outperforming existing nighttime image dehazing methods by a substantial margin of 15.5\% for MUSIQ  and 23.5\% for ClipIQA. 
\end{abstract}

%

\begin{figure*}[t!]
    \centering
    \includegraphics[width=\linewidth, height = 0.60\linewidth]{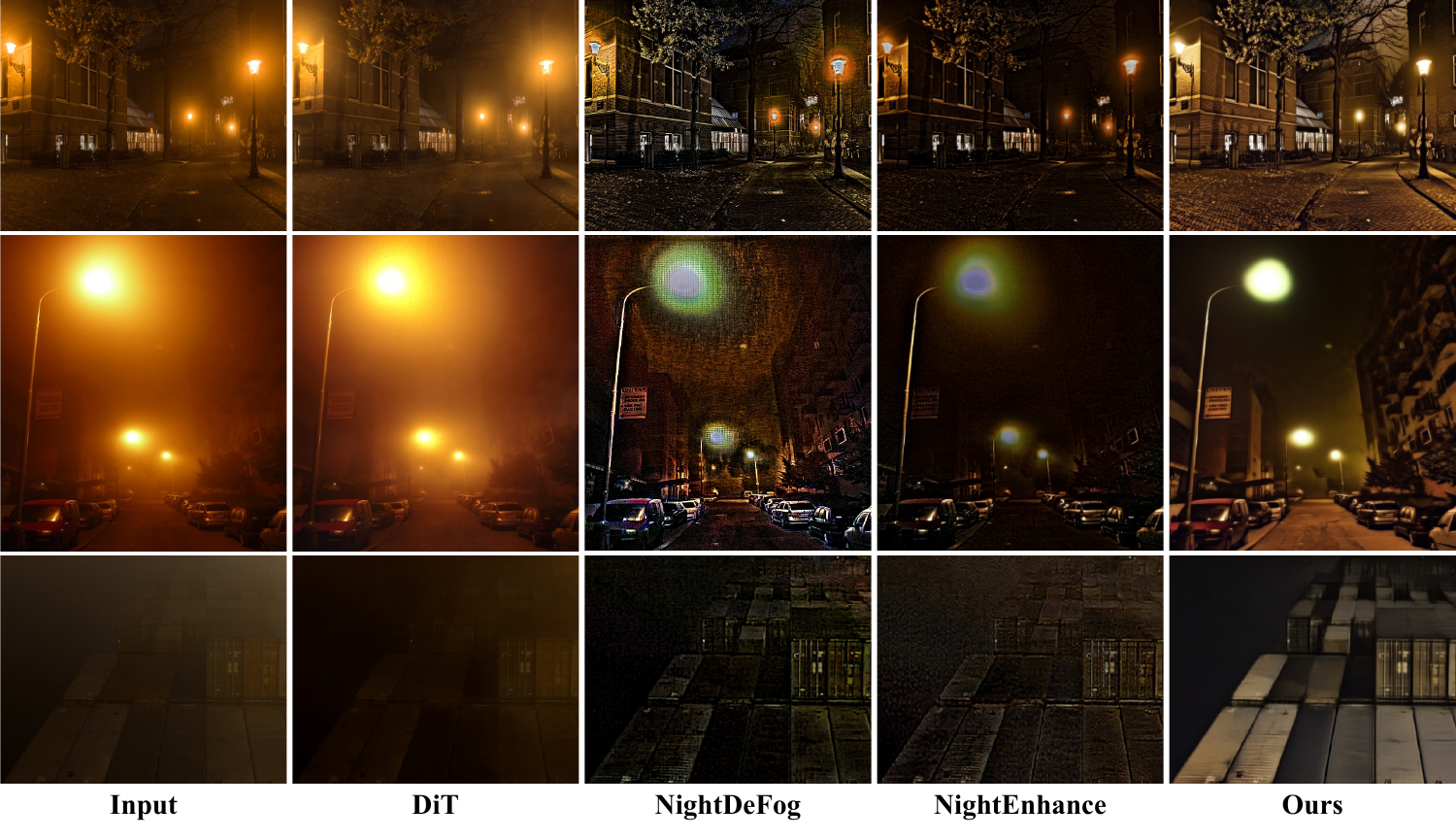}
    \captionof{figure}{Qualitative results from NightEnhance'23~\cite{jin2023enhancing}, NightDeFog'20~\cite{yan2020nighttime}, DiT'23~\cite{Peebles2023DiT} and our method, on the real-world dataset.
        Ours not only suppress glow but also reveal the detailed textures of the night scenes, including those under low light and strong glow.
    }
    
    \label{fig_trailer}
\end{figure*}%

\section{Introduction}
		
Dehazing night images deals not only with haze but also with nighttime degradations such as low light, noise, glow, uneven light distribution, etc. The combination of these degradations and the absence of paired real data make the task significantly challenging.

Existing supervised and semi-supervised nighttime image dehazing methods~\cite{yan2020nighttime, jin2023enhancing, liu2022nighttime} use either synthetic data or both synthetic and real data in training. Unfortunately, as shown in Figure~\ref{fig_trailer}, their performance is unsatisfactory when evaluated on real-world nighttime haze images. 
The main reason is that there is a large domain gap between synthetic haze and real-world haze, and unpaired real data cannot provide pixel-level guidance. As a result, models trained on these data lead to inferior performance when applied to real haze images.
Table \ref{tab_quan} shows five non-reference image quality assessment (NF-IQA) scores of these methods on real night haze data. Noticeably, there are significant gaps between the NF-IQA scores of clear night images and those of restored results obtained from existing methods.

Masked autoencoders (MAE)~\cite{he2022masked} are known to yield strong representations for high-level tasks. The core design of MAE is to use severe augmentation for training. MAE masks 75\% of the input image before reconstructing it. 
This paper aims to bring MAE-like strong representations to nighttime image dehazing. However, MAE's masking strategies are designed to capture structural spatial priors by filling in the masked regions. In contrast, nighttime image dehazing aims to restore a clear background behind haze and nighttime degradations.

To obtain strong priors that are resilient to real-world night haze degradations, we design the severe augmentation for our learning process. Our main novelty lies in defining ``severe” based on proper augmentation. Specifically, we first find two key challenging factors of nighttime images for augmentation: light effects and noise. During training, we intentionally degrade clear nighttime images by blending them with various light effects, including glow, and adding noise. 
As demonstrated in MAE~\cite{he2022masked}, the severity of augmentation is key to yield strong feature representations or priors. Accordingly, we define ``severe augmentation” as noise values approaching as high as the pixel intensity values blended with strong glow and other night light effects.

Based on our severe augmentation, we propose NightHaze, a simple yet effective nighttime image dehazing method with self-prior learning. 
Our main novelty, the design of severe augmentation, allows our network to learn the strong priors that are robust to real-world haze effects. Once the strong priors are obtained, our network can effectively enhance the visibility of nighttime haze images. Like MAE, we apply severe augmentation and then train our network to restore the clear background. Table \ref{tab_quan} shows that the NF-IQA scores of our self-prior learning are closer to the NF-IQA scores of clear night images.

While our self-prior learning effectively can enhance the visibility of night haze images, in some cases, artifacts can persist, typically in the form of over-suppression.
To refine the output of our self-prior learning, we introduce a self-refinement module based on the teacher-student framework~\cite{tarvainen2017mean}. 
Unlike standard teacher-student frameworks that directly pass knowledge from the student to the teacher, we utilize NF-IQA scores to monitor the quality of knowledge from the student. This allows us to pass only high-quality information to the teacher, resulting in superior performance.
Figure \ref{fig_trailer} shows our results in comparison to those of the state-of-the-art methods.
The main contributions are as follows:
\begin{itemize}
    \item 
    We propose self-prior learning to bring MAE-like strong representations to nighttime image dehazing. Our main novelty lies in the design of severe augmentation, which enables our model to learn clear background priors.
    \item  
    We propose self-refinement to refine our model by utilizing real-world nighttime haze images. Unlike the standard teacher-student framework, we introduce NF-IQA scores to assess the knowledge quality from our student model's learning. This approach avoids inaccurate knowledge transfer from our student to our teacher.
    \item 
    Extensive experiments demonstrate that our Nighthaze achieves a significant performance improvement on real-world nighttime haze images. Our NightHaze attains outstanding scores of 52.87 in MUSIQ and 76.88 in ClipIQA, significantly exceeding existing nighttime image dehazing methods with a substantial margin of 15.5\% for MUSIQ and 23.5\% for ClipIQA.
\end{itemize}



\section{Related Work}
\label{sec_related}
\paragraph{\textbf{Nighttime Image Dehazing}}
Previous image dehazing methods~\cite{tan2008visibility, fattal2008single, he2011single,berman2018single} remove haze effects based on the haze model. Specifically, they estimate transmission maps and atmosphere lights to recover the clear background. However, nighttime haze images always suffer from noise. Thus, these methods are hard to estimate accurate transmission maps and atmosphere lights, resulting in inferior performance. 
Recent deep learning-based daytime image dehazing methods~\cite{ye2022perceiving,yu2022frequency, dong2020multi, qin2020ffa, zheng2021ultra, wu2021contrastive, guo2022image,song2023vision, shao2020domain, chen2021psd, li2022physically, li2020zero, li2021you, huang2019towards, liu2020end, zhao2021refinednet} propose different networks to restore haze images. These methods can address nighttime haze images by training on synthetic nighttime dehazing datasets. 
However, these methods struggle to handle nighttime haze images as they often neglect complex nighttime degradation factors.

Existing optimization-based nighttime dehazing methods \cite{wang2022variational,ancuti2016night,ancuti2020day, zhang2014nighttime,zhang2017fast,tang2021nighttime,liu2022nighttime} address nighttime haze effects using the atmospheric scattering model, new imaging model, among others. 
However, these methods struggle to handle challenging hazy regions affected by significant noise and complex lighting conditions. The main issue is the difficulty in estimating atmospheric lights in these regions.

Existing learning-based nighttime dehazing methods~\cite{zhang2020nighttime,yan2020nighttime,kuanar2022multi, jin2023enhancing} propose CNNs for nighttime image dehazing. 
However, the synthetic datasets they used were constructed using less challenging augmentations, such as regions affected by light noise. Models trained using these datasets struggle to restore clear backgrounds in challenging real-world haze images, including those affected by significant noise and complex lighting conditions. Unsupervised methods ~\cite{yan2020nighttime,jin2023enhancing}, which are based on unpaired real-world data, cannot provide pixel-level guidance for training.

\paragraph{\textbf{Image Quality Assessment}} To qualitatively evaluate the performance of real-world haze images without the corresponding ground truth, we introduce several image quality assessment methods, including MUSIQ~\cite{ke2021musiq}, ManIQA~\cite{yang2022maniqa}, ClipIQA~\cite{wang2023exploring}, HyperIQA~\cite{su2020blindly}, and TRES~\cite{golestaneh2022no}. MUSIQ~\cite{ke2021musiq} is builted based on a transofmer network and evaluate the image quality at multiple image scales. ManIQA~\cite{yang2022maniqa} integrates the global and local information of images to assess the image quality. HyperIQA~\cite{su2020blindly} assesses in-the-wild image quality by using a self-adaptive hyper network architecture. TRES~\cite{golestaneh2022no} leverages transformers, relative ranking, and self-consistency losses to effectively evaluate the image quality. ClipIQA~\cite{wang2023exploring} introduces Contrastive Language-Image Pre-training (CLIP) models to evaluate the quality perception and abstract perception of images.

\begin{figure*}[t!]
    \centering
    \includegraphics[width=1\linewidth]{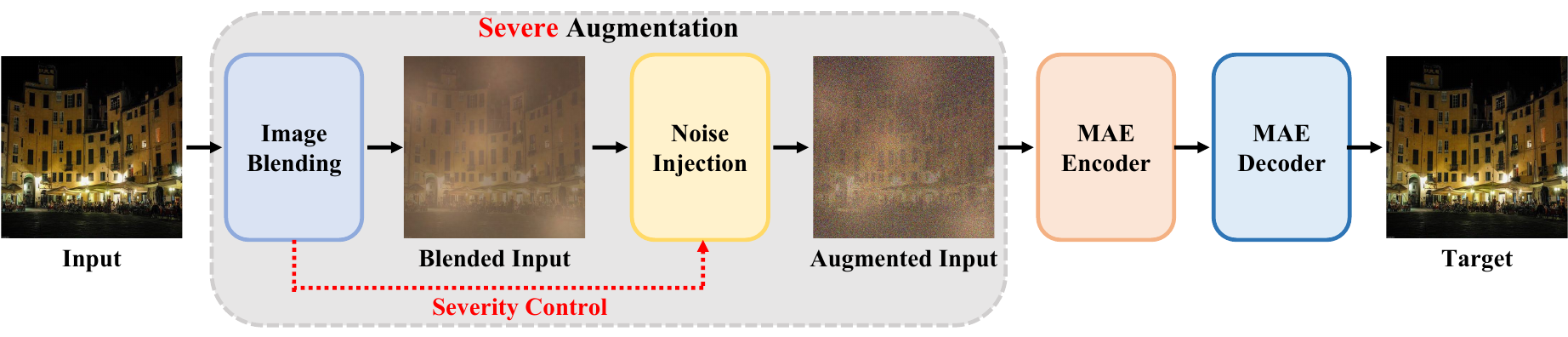}
    \caption{Overview of our self-prior learning. This approach aims to produce strong priors that are resilient to real-world night haze degradations. Given a clear nighttime image, we blend clear images with various maps collected from real-world haze images and add noise to these blended images. The augmented input is then fed into an Encoder-Decoder framework to recover the clear background. The loss is a simple L1 loss between the reconstructed image and the clear input. 
    The key to self-prior learning is the severity of the augmentation. By increasing the noise values to approach as high as the pixel intensity values of the glow and light effect blended images, our augmentation becomes severe, resulting in stronger priors.
    }
    \label{overview}
\end{figure*}

\begin{figure}[t]
    \centering
    {\includegraphics[width=4cm, height=1.5cm]{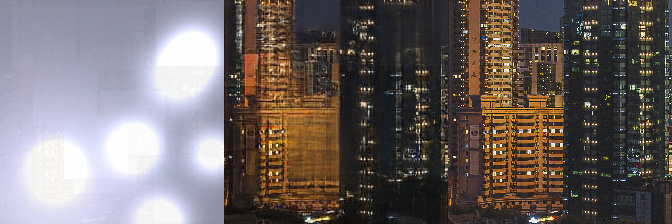}}\hspace{0.3pt}
    {\includegraphics[width=4cm, height=1.5cm]{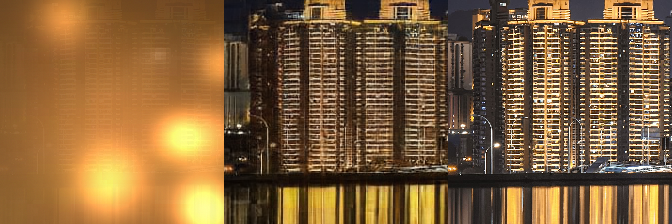}}\hspace{0.3pt}
    \\
    {\includegraphics[width=4cm, height=1.5cm]{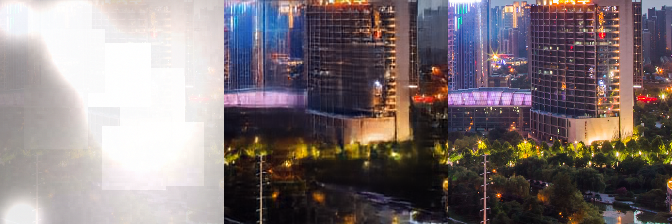}}\hspace{0.3pt}
    {\includegraphics[width=4cm, height=1.5cm]{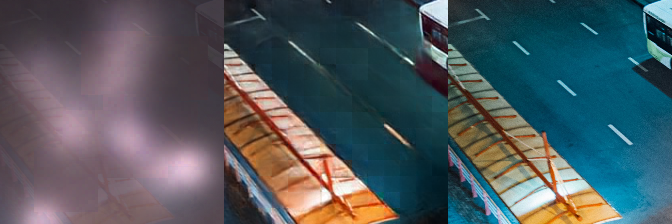}}\hspace{0.3pt}
        \\
    {\includegraphics[width=4cm, height=1.5cm]{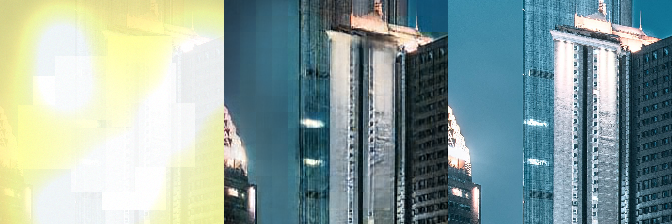}}\hspace{0.3pt}
    {\includegraphics[width=4cm, height=1.5cm]{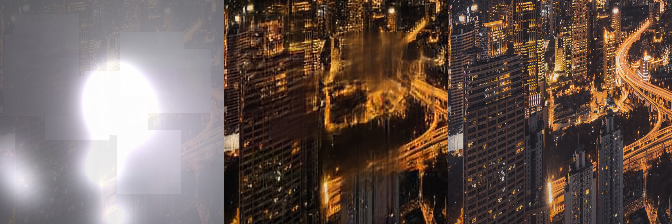}}\hspace{0.3pt}
    \\
    \caption{Example results on validation sets. For each triplet, we show the augmented image (left), our restoration (middle), and the ground-truth (right).
    }
    \label{fig_train}
\end{figure}

\section{Proposed Method}

The overview of our NightHaze is shown in Figure~\ref{overview}. It includes two components: self-prior learning and self-refinement. 
Our self-prior learning aims to bring the MAE-like framework to nighttime image enhancement, resulting in strong network priors that are resilient to real-world night haze effects.
%
Our self-refinement utilizes unlabeled nighttime haze images to refine the dehazing ability of our network in the real scenes in a semi-supervised manner.

\subsection{Self-Prior Learning}
\label{method_spl}
Our self-prior learning aims to learn strong network priors of clear backgrounds using severe augmentation.
As shown in Figure~\ref{overview}, our training process is similar to the approach in MAE, where we severely augment clear images and force the network to recover their clear backgrounds. 
MAE employs masking strategies to augment the input images. This masking forces the network to learn the structural spatial priors of the world by predicting the contents in the masked regions.
However, nighttime image dehazing is about restoring a clear background behind haze and nighttime degradations. Hence, masking is not a suitable augmentation for dehazing.

Nighttime haze degradations are mainly caused by glow light effects and noise. Due to this, we degrade clear images by blending them with light effects including glow and adding noise and then we restore the clear images, as demonstrated in Figure~\ref{fig_train}. The augmentation can be formulated as:
\begin{equation}
    \label{eqn_generate_model}
    I = W_{b}*J + (1-W_{b})*L + \epsilon,
\end{equation}
where I is the augmented image, J is the clear image, $W_{b}$ is the blend weight map, $L$ is the light map,  $\epsilon$ is the noise. We explain the details of each term in Eq.~\eqref{eqn_generate_model} as follows.

\begin{figure}[t]
    \centering
    {\includegraphics[width=2.0cm, height=2.0cm]{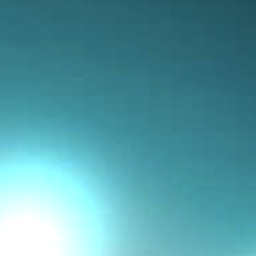}}\hspace{0.5pt}
    {\includegraphics[width=2.0cm, height=2.0cm]{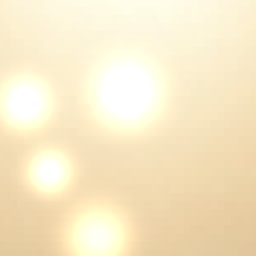}}\hspace{0.5pt}
    {\includegraphics[width=2.0cm, height=2.0cm]{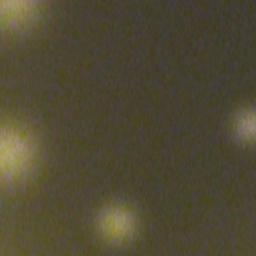}}\hspace{0.5pt}
    {\includegraphics[width=2.0cm, height=2.0cm]{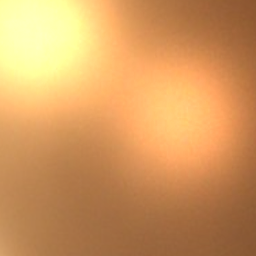}}\hspace{0.5pt}
    \caption{Visualization of different light maps that contain glow effects. We use Gaussian kernels to simulate glow effects. By adjusting the parameters of kernels, we can control the number, size, and brightness of glow regions.
    }
    \label{fig_augA}
\end{figure}

\noindent{\textbf{Light Map $L$}} We manually capture atmosphere light maps from real haze images. When regions in a night image suffer from dense haze effects (where the signal is less than the noise), they can be regarded as atmospheric lights. Hence, we collect these regions from real haze images as light maps. 
Given an input image $J_{i} \in \mathbb{R}^{C_{i} \times H_{i} \times W_{i}}$, where $C_{i}$ is the number of channels and ($H_{i}, W_{i}$) is image size, we first sample a light map and resize it to $\mathbb{R}^{C_{i} \times H_{i} \times W_{i}}$.
Moreover, to simulate glow effects, we use Gaussian kernels to enhance the brightness of several random regions in the light maps. Figure~\ref{fig_augA} shows our augmented light maps.

\begin{figure}[t]
    \centering
    \subfloat[$t$ = 0.2]
    {\includegraphics[width=2cm, height=2cm]{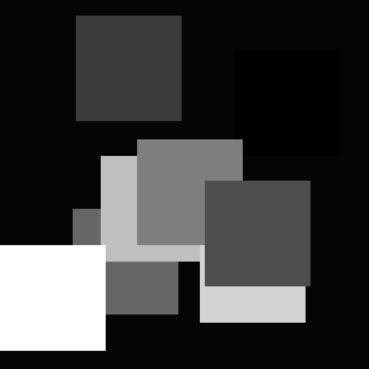}}\hspace{0.5pt}
    \subfloat[$t$ = 0.4]
    {\includegraphics[width=2cm, height=2cm]{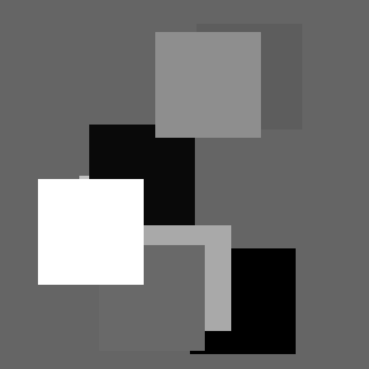}}\hspace{0.5pt}
    \subfloat[$t$ = 0.6]
    {\includegraphics[width=2cm, height=2cm]{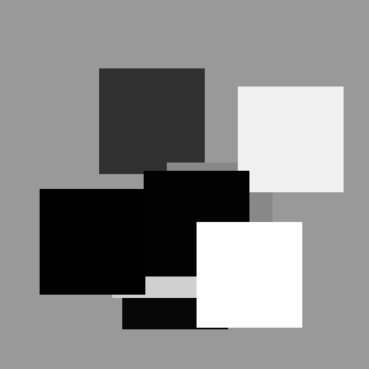}}\hspace{0.5pt}
    \subfloat[$t$ = 0.8]
    {\includegraphics[width=2cm, height=2cm]{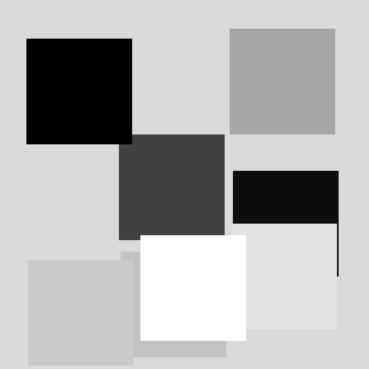}}\hspace{0.5pt}
    \caption{Visualization of blending weight maps, where the black and white regions represent low and high blending values. By adjusting the values of blending weights and the noise term, we can control the severity of our augmentation.
    }
    \label{fig_trans}
\end{figure}

\noindent{\textbf{Blending Weight Map $W_{b}$}} As the goal of our blending weight $W_{b}$ is to mix a clear image and light maps, the range of the blending weight $W_{b}$ is (0, 1). To obtain the blending weight $W_{b}$, we first sample a value $t \in \mathbb{R}^{1}$ from a uniform distribution in [$T_{\rm Low}$, $T_{\rm High}$], where both $T_{\rm Low}$ and $T_{\rm High}$ are lower than 1. We then expand $t \in \mathbb{R}^{1}$ to a map $W_{b} \in \mathbb{R}^{C_{i} \times H_{i} \times W_{i}}$. Moreover, we randomly adjust some regions' blending weights. Figure~\ref{fig_trans} visualizes several blending weight maps. Our blending weight map is inspired by transmission maps of the haze formulation. However, transmission maps rely on the depth of images for their generation. Obtaining accurate depth maps in nighttime images is a challenging problem. A straightforward approach is to use monocular depth estimates, but this may introduce errors. Unlike transmission maps, we randomly generate uneven blending weights for an image. This approach enables us to simply and effectively augment clear images.

\noindent{\textbf{Noise $\epsilon$}} We take Gaussian noise as the noise $\epsilon$, which can be formulated as $W_{\epsilon} * \mathcal{N}(0,1)$, where $W_{\epsilon}$ is the weight of the Gaussian noise. Theoretically, $\mathcal{N}(0,1)$ ranges from negative infinity to positive infinity, with about approximately 99.7\% of the values fall within the range of -3 to 3. We remove the rest 0.3\% values and the negative values of $\mathcal{N}(0,1)$. Thus, the range of our noise term $\epsilon$ is (0, 3*$W_{\epsilon}$).

\noindent{\textbf{Severe Augmentation}} MAE demonstrates that severe augmentation is key to obtaining strong representations. Motivated by this, we define ``severe augmentation” as $\epsilon$ approaching $W_{b}*J$ according to Eq.~\eqref{eqn_generate_model}. 
The main reason is that the level of the noise $\epsilon$ affects the difficulty of nighttime image restoration. 
To be more specific, we use min-max normalization to normalize the clear image and thus the range of $J$ is (0, 1). After adding light effects, the range of the term $W_{b}*J$ becomes to (0, $T_{\rm High}$). Since its range depend on the value of $T_{\rm High}$, we further define $W_{\epsilon}$ as $W_{n} * T_{\rm High}$ and thus the range of $\epsilon$ becomes to (0, 3*$W_{n}$*$T_{\rm High}$). We empirically find that when $W_{n} = 0.1$ and $\epsilon \in [0, 0.3*T_{\rm High}]$, our augmentation is difficult enough.

In light maps, we use Gaussian kernels to increase the brightness of several random regions. These regions can overlap, resulting in uneven light distribution in the augmented light maps. Moreover, due to the overlapping brightness enhancements, the glow effects can become oversaturated. When blending clear images with these oversaturated light effects, their signals can dominate, making restoration challenging.  

\begin{table*}[ht!]
    \centering
  \renewcommand\arraystretch{1.2} 
   \fontsize{9}{10}\selectfont
        \begin{tabular}{l|c|c|c|c|c|c}
            \toprule
            Non-reference metrics & MUSIQ & TRES  & Hyperiqa & CLIPIQA & MANIQA & Contrast \\
            \midrule
            Statistics from clear images & {\textbf{53.91}} & {\textbf{78.69}} & {\textbf{0.6819}} & {\textbf{0.6834}} & {\textbf{0.4319}}  & {\textbf{50.96}} \\
            \hline
            Statistics from haze images &   {\textbf{42.12}} & {\textbf{57.94}} &	{\textbf{0.5186}} & 	{\textbf{0.5267}} &	{\textbf{0.2556}} &	{\textbf{22.21}} \\
            \hline
            \hline
            Backbone with syn-training & 44.25 &	60.84 &	0.5903 &	0.5449 & 	0.2665  & 24.08 \\
            \hline
            \hline
            NightDeFog~\cite{yan2020nighttime} & 45.76	& 65.83 &	0.6133 &	0.5146 &	0.3318 & 	57.60  \\
            \hline
            NightVDM~\cite{liu2022nighttime} & 40.90 & 	53.87	& 0.5241 & 	0.4656	& 0.2332 & 	27.47    \\
            \hline
            Restormer~\cite{zamir2022restormer}  & 39.13 &	52.97 &	0.4956 & 	0.4763 &	0.2428  &	24.58  \\
            \hline
            Uformer~\cite{wang2022uformer} & 39.14 &	52.60 &	0.4902 &	0.4427 &	0.2420  &	25.03  \\
            \hline
            WeatherDiffusion~\cite{ozdenizci2023restoring} & 38.98 &	52.33 &	0.4868 &	0.5044 &	0.2338 &	24.04  \\
            \hline
            NightEnhance~\cite{jin2023enhancing} &   44.33	& 60.07 &	0.5379 &	0.4736 &	0.2354 &	24.57    \\
            \hline
            DiT~\cite{Peebles2023DiT} & 41.02  &	55.54  &	0.5259  &	0.5059 &	0.2513 	 & 23.94  \\
            \hline
            Ours & {\textbf{52.87}} & 	{\textbf{76.88}} & 	{\textbf{0.6747}} & 	{\textbf{0.6360}} & 	{\textbf{0.4009}} & 	{\textbf{44.35}}  \\
            \bottomrule
        \end{tabular}%
    \caption{Quantitative comparison on the RealNightHaze dataset. For non-reference metrics, higher is better. In terms of image contrast, the better it is when the contrast closely matches the statistical value of a clear image. NightDeFog, NightVDM and NightEnhance are nighttime image dehazing methods, while the rest methods are image restoration backbones. ``syn-training” means training our backbone on synthetic datasets.}
    \label{tab_quan}%
\end{table*}%

\noindent{\textbf{Implementation}} Given a clear nighttime image, we first sample a light map $L$ from our dataset and randomly enhance the brightness of several regions. 
Second, we sample a value $t$ from a uniform distribution to generate a uniform map. We then randomly adjust the values of several regions in this uniform map to create the blending weight map $W_{b}$.
Third, we generate a significant Gaussian noise map $\epsilon$. 
Finally, we augment the input clear image using Eq.~\eqref{eqn_generate_model}, as illustrated in Figure \ref{fig_train} which shows several augmented inputs. During the training stage, we calculate the L1 loss between the reconstructed and original clear images.

\subsection{Self-Refinement}
\label{method_srf} 
Although our self-prior learning effectively handles haze effects, it occasionally results in artifacts, particularly in the form of over-suppression. 
To address this issue, we propose a teacher-student framework \cite{lin2024nightrain, Chen2024Dual} to further refine our network using real haze images. The network parameter from our self-prior learning is used to initialize a teacher model $w_{T}$ and a student model $w_{S}$. The purpose of the teacher model is to generate the pseudo ground truth based on confidence score, while the student model is trained on high-confidence predictions and their corresponding inputs. By gradually learning from high-confidence predictions, our network is allowed to address more challenging regions.

Given a real-world hazy dataset $\mathbf{D}_{\rm uh} = \{ \mathbf{x}^{\rm uh}_{i}\}^{N_{\rm uh}}_{i=1}$, where $\mathbf{x}^{\rm uh}_{i}$ is the $i$-th haze image and $N_{\rm uh}$ is the number of haze images, we utilize the teacher model $w_{\rm T}$ to generate predictions $\mathbf{y}^{\rm uh}_{i}$ and confidence maps $\mathbf{u}^{\rm uh}_{i}$. Specifically, we first sample overlapping regions from $\mathbf{x}^{\rm uh}_{i}$ and input the set of overlapping regions into our teacher model $w_{T}$ to generate predictions. Then, we calculate the mean values and variance of overlapping regions as predictions and confidence maps. Subsequently, we use a threshold $\mathbf{v}^1_{thr}$ to convert the confidence map $\mathbf{u}^{\rm uh}_{i}$ to a binary mask $\mathbf{m}^{\rm uh}_{i}$.

During the training stage, we use the noise and haze augmentation to degrade the haze images $\mathbf{x}^{\rm uh}_{i}$. The degraded haze images $\mathbf{\hat{x}}^{\rm uh}_{i}$ are then fed into our student model for training. The loss function can be formulated as:
\begin{equation}
    \label{eqn_loss}
    {\rm Loss} = \frac{1}{N} \sum_{i=1}^{N} |\mathbf{\hat{y}}^{\rm uh}_{i}- \mathbf{y}^{\rm uh}_{i}| * \mathbf{m}^{\rm uh}_{i},
\end{equation}
where $N$ is the number of samples in the training batch and $\mathbf{\hat{y}}^{\rm uh}_{i}$ is the output of the degraded haze images $\mathbf{\hat{x}}^{\rm uh}_{i}$. Once the losses are obtained, traditional teacher-student frameworks use back-propagation to update the student model $w_{S}$. Then they use the Exponential Moving Average (EMA) to update the teacher model. 
However, the student model is trained using augmented real-world haze inputs. Since the augmentation is random, the updates of the student model in some training steps may be wrong and thus affect the performance of the teacher model by EMA. To avoid this, we introduce a non-reference metric to monitor each update of the student model, which can be formulated as:

\begin{equation}
    \label{eqn_check}
    {\rm Score} = F_{\rm IQA}(F(w^{t+1}_{s}, \mathbf{x}^{\rm uh}_{i})) + F_{\rm IQA}(F(w^{t}_{s}, \mathbf{x}^{\rm uh}_{i})),
\end{equation}
where $F_{\rm IQA}(\cdot)$ represents non-reference image assessment metrics, $F(w^{t+1}_{s}, \mathbf{x}^{\rm uh}_{i})$ and $F(w^{t}_{s}, \mathbf{x}^{\rm uh}_{i})$ are $\mathbf{x}^{\rm uh}_{i}$'s outputs generated using the student weights $w^{t+1}_{s}$ and $w^{t}_{s}$. $w^{t}_{s}$ is the current weight of the student model, while $w^{t+1}_{s}$ is the updated weight. If Score $ > \mathbf{v}^2_{thr}$, the student model will accept this update and pass the knowledge to the teacher model. Otherwise, the student model will not accept this update.

\section{Experiments}

Due to the lack of ground truth, previous methods typically use synthetic datasets for quantitative evaluation and real-world datasets for qualitative evaluation. 
However, the quantitative evaluation on synthetic datasets cannot effectively represent the real-world dehazing capabilities. Motivated by this, we introduce a new benchmark to quantitatively evaluate the dehazing performance on real-world nighttime haze datasets. 
Specifically, we use five non-reference image quality assessments, including MUSIQ~\cite{ke2021musiq}, ManIQA~\cite{yang2022maniqa}, ClipIQA~\cite{wang2023exploring}, HyperIQA~\cite{su2020blindly}, and TRES~\cite{golestaneh2022no}, along with image contrast, to evaluate the performance of existing methods.

Our real-world nighttime haze datasets, \textit{RealNightHaze}, is collected from internet and existing nighttime dehazing methods \cite{zhang2020nighttime, jin2023enhancing}, and consists of 440 real-world nighttime haze images.

\begin{figure*}[t!]
    \centering
    {\includegraphics[width=3.0cm, height=3.0cm]{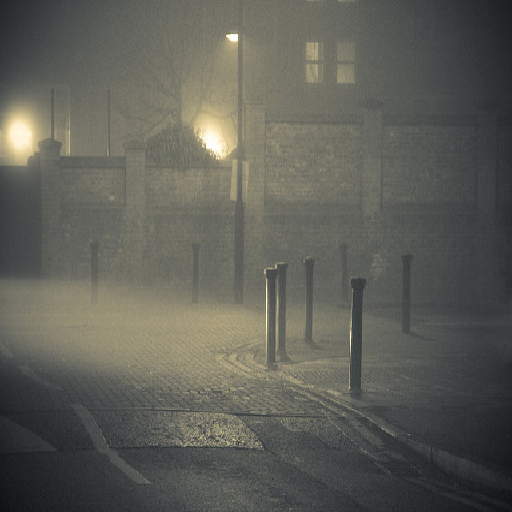}}\hspace{1pt}
    {\includegraphics[width=3.0cm, height=3.0cm]{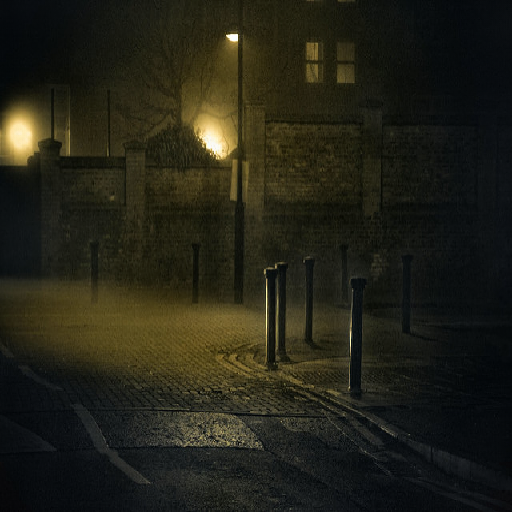}}\hspace{1pt}
    {\includegraphics[width=3.0cm, height=3.0cm]{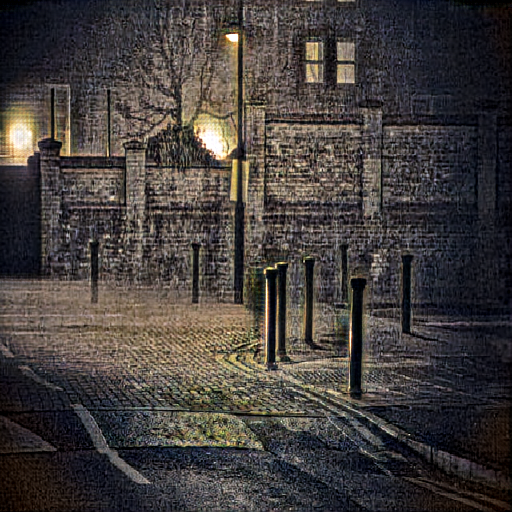}}\hspace{1pt}
    {\includegraphics[width=3.0cm, height=3.0cm]{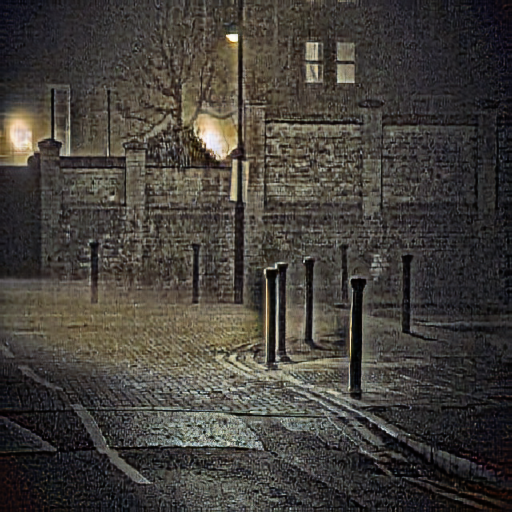}}\hspace{1pt}
    {\includegraphics[width=3.0cm, height=3.0cm]{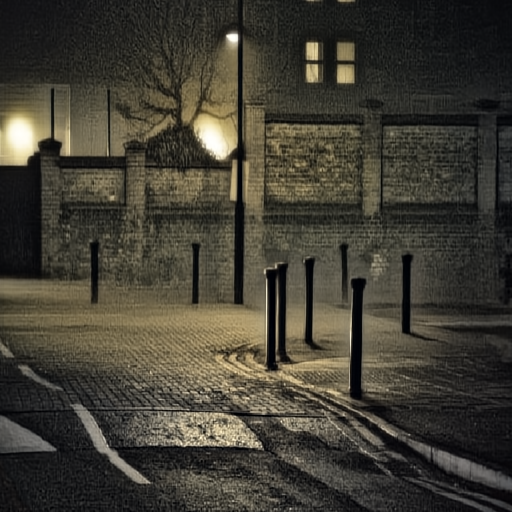}}\hspace{1pt}
    \subfloat[Inputs]
    {\includegraphics[width=3.0cm, height=3.0cm]{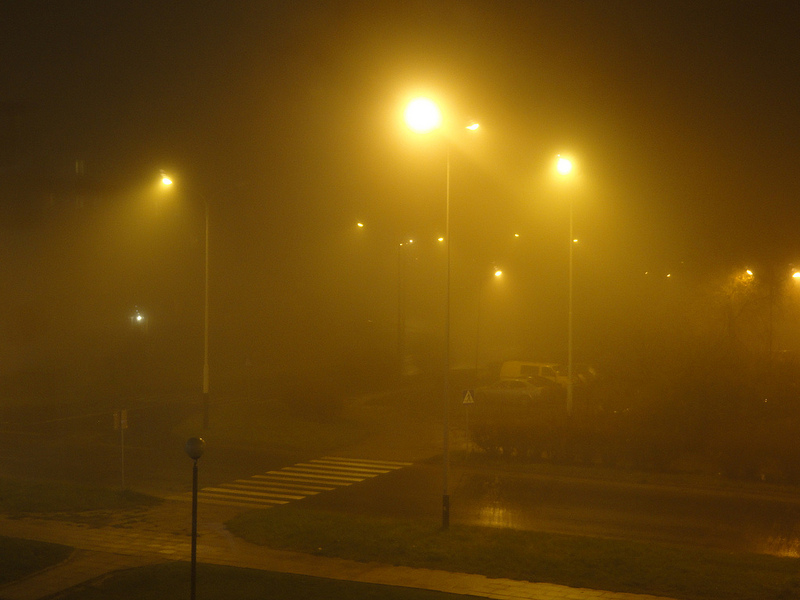}}\hspace{1pt}
    \subfloat[DiT]
    {\includegraphics[width=3.0cm, height=3.0cm]{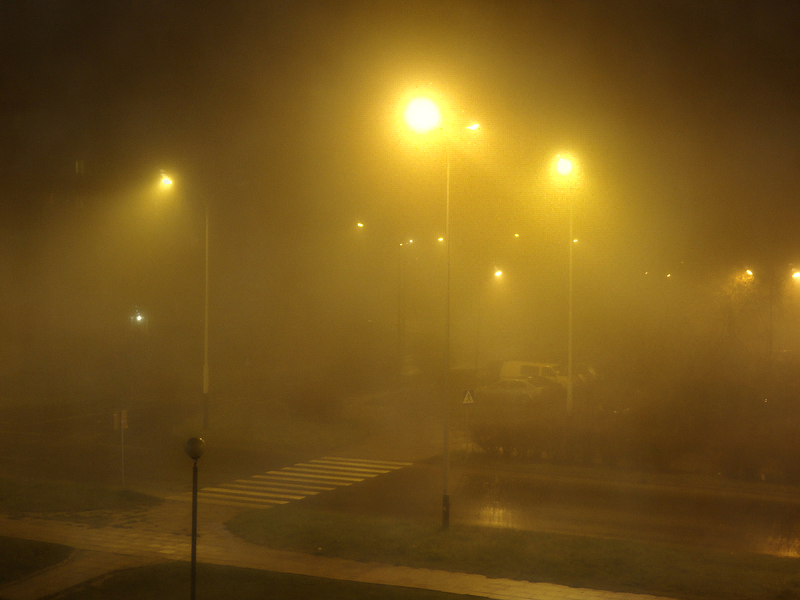}}\hspace{1pt}
    \subfloat[NightDeFog]
    {\includegraphics[width=3.0cm, height=3.0cm]{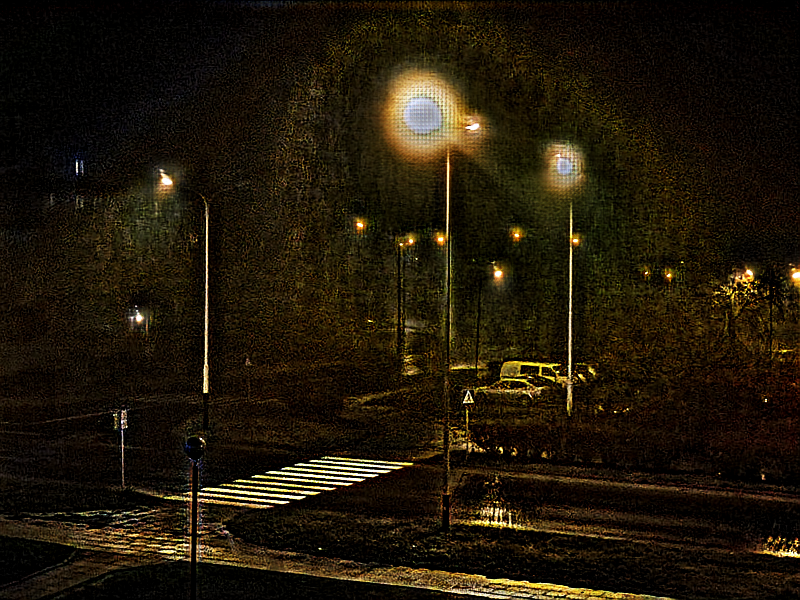}}\hspace{1pt}
    \subfloat[NightEnhance]
    {\includegraphics[width=3.0cm, height=3.0cm]{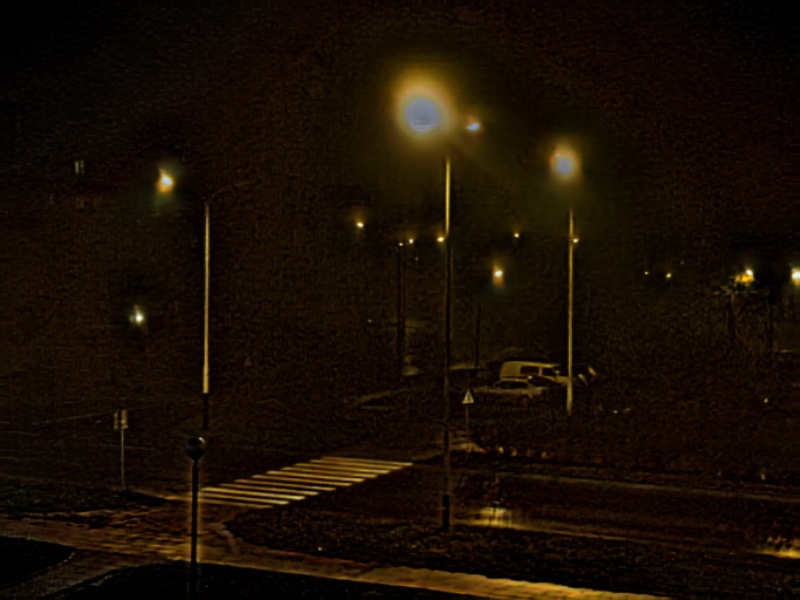}}\hspace{1pt}
    \subfloat[\textbf{Ours}]
    {\includegraphics[width=3.0cm, height=3.0cm]{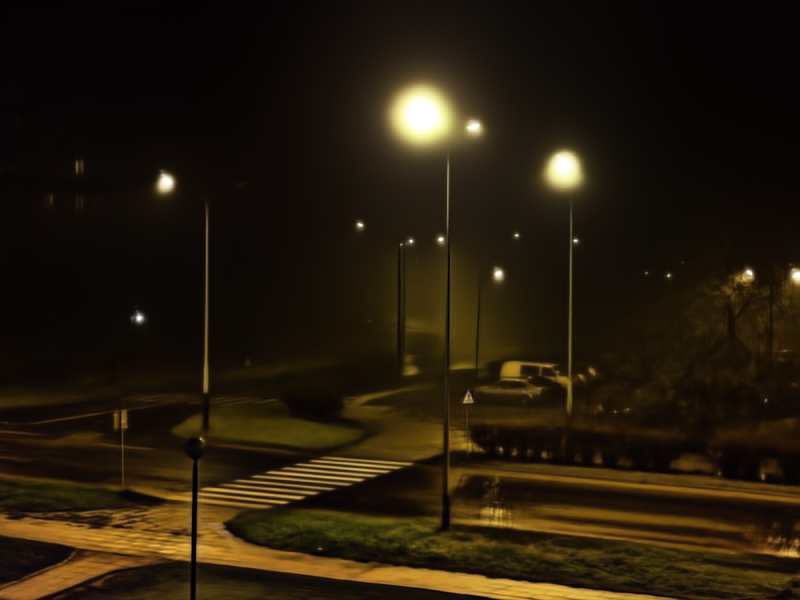}}\hspace{1pt}

    \caption{Qualitative results from NightEnhance'23~\cite{jin2023enhancing}, NightDeFog'20~\cite{yan2020nighttime}, DiT'23~\cite{Peebles2023DiT}, and our method, on the real-world dataset.
    Our method not only handles glow effects but also reveals detailed textures in low-light regions, such as walls and streets. 
    Zoom in for better visualisation.
    }
    \label{fig_exp1}
\end{figure*}

\subsection{Implementation Details}
Our NightHaze includes two parts: self-prior learning and self-refinement. We adopt the MAE network \cite{he2022masked} as our backbone. 
In our self-prior learning, the size of input images is 224 $\times$ 224. 
During the training stage, we use Adam as our optimizer and set the initial learning rate to 1.5e-4. 
The total training steps and the training batch size are set to 20,000 and 128, respectively.
Our self-prior learning consists of three parts: the light map, the blending weight map, and noise.

\noindent{\textbf{Light Map}} At each training step, we randomly select an atmospheric light map for every clear image. Then, we select various regions and amplify their brightness using Gaussian kernels. The number of selected regions is between 2 and 10, and the size of Gaussian kernels ranges from 15 to 80. 

\noindent{\textbf{Blending Weight Map}} 
We initially select a blending weight $t$ from a uniform distribution ($T_{\rm low}$, $T_{\rm high}$) and expand it into a map.
Then, we randomly adjust the blending weights of several regions. $T_{\rm low}$ and $T_{\rm high}$ are set to 0.001 and 0.1, respectively. We set the number of the selected regions to 8. The size of each region is 64 $\times$ 64 and the range of the changed value is a uniform distribution (0, 0.04).

\noindent{\textbf{Noise}} During the training stage, we use Gaussian noise to augment our input images. We set $W_{n}$ to 0.1. Thus, the value range of our $\epsilon$ becomes to (0, 0.03).

\noindent{\textbf{Self-Refinement}} 
For a haze image of size 256 $\times$ 256, we first sample overlapping regions within the input haze image.
The image size of the overlapping regions is 224 $\times$ 224 and the stride of the overlap sampling is 4. 
Once the pseudo ground truths are obtained, we randomly crop 224 $\times$ 224 regions from both the haze images and the pseudo ground truths for training our student model.
We use Adam as our optimizer, with the initial learning rate set to 2e-5. We set the training step and the training batch size to 10,000 and 16, respectively. The EMA weight is set to 0.9999. The thresholds $\mathbf{v}^1_{thr}$ and $\mathbf{v}^2_{thr}$ are set to 0.005 and 0.

\begin{table*}[t]
    \centering

  \renewcommand\arraystretch{1.2} 
   \fontsize{9}{10}\selectfont

  \setlength{\tabcolsep}{4.0mm}{
        \begin{tabular}{l|c|c|c|c|c|c}
            \toprule
            Non-reference metrics & MUSIQ & TRES  & Hyperiqa & CLIPIQA & MANIQA & Contrast \\
            \midrule
            Statistics from clear images & {\textbf{53.91}} & {\textbf{78.69}} & {\textbf{0.6819}} & {\textbf{0.6834}} & {\textbf{0.4319}}  & {\textbf{50.96}} \\
            \hline
            Statistics from haze images &   {\textbf{42.12}} & {\textbf{57.94}} &	{\textbf{0.5186}} & 	{\textbf{0.5267}} &	{\textbf{0.2556}} &	{\textbf{22.21}} \\
            \hline
            \hline
            Backbone with syn-training & 44.25 &	60.84 &	0.5903 &	0.5449 & 	0.2665  & 24.08 \\
            \hline
            \hline
            Backbone with 0\% severe & 44.67 &	62.18 &	0.5773 &	0.5364 &	0.2930 &	23.40 \\
            \hline
            Backbone with 25\% severe &  48.64 &	67.96 &	0.6253 &	0.5715 &	0.3361 & 28.73  \\
            \hline
            Backbone with 50\% severe & 49.12 &	68.74 &	0.6286 &	0.5750 &	0.3417 &	29.80  \\
            \hline
            Backbone with 75\% severe & 49.31 &	68.92 &	0.6289 &	0.5891& 	0.3420 &	31.15  \\
            \hline
            Backbone with 100\% severe & {\textbf{52.87}} & 	{\textbf{76.88}} & 	{\textbf{0.6747}} & 	{\textbf{0.6360}} & 	{\textbf{0.4009}} & 	{\textbf{44.35}}  \\
            \bottomrule
        \end{tabular}%
    }
    \caption{Ablation studies on the RealNightHaze dataset. For non-reference metrics, higher is better. In terms of image contrast, the better it is when the contrast closely matches the statistical value of a clear image. ``xx\% severe” means xx\% of the training data was severely augmented, while the remaining data was non-severely augmented. ``syn-training” means training our backbone on synthetic datasets.
    }
    \label{tab_abl}%
\end{table*}%

\subsection{Quantitative Evaluation on RealNightHaze}
We compare our NightHaze method on RealNightHaze with state-of-the-art (SOTA) methods, including NightEnhance~\cite{jin2023enhancing}, NightDeFog~\cite{yan2020nighttime}, NightVDM~\cite{liu2022nighttime}, Uformer~\cite{wang2022uformer}, Restormer~\cite{zamir2022restormer}, WeatherDiffusion~\cite{ozdenizci2023restoring} and DiT~\cite{Peebles2023DiT}. Table~\ref{tab_quan} shows that our method achieves a significant performance improvement in all non-reference metrics. 
For example, the MUSIQ and TRES of NightDeFog are 45.76 and 65.83, respectively. Our method achieves 52.87 of MUSIQ and 76.88 of TRES, which outperforms NightDeFog by 7.11 and 11.05, respectively. Furthermore, it can be observed that our image quality scores are closer to those of clean night images. 
This is because we propose self-prior learning, which brings MAE-like strong representations to nighttime image enhancement. 
By utilizing severe augmentation, we can generate strong priors that demonstrate robustness against real-world haze effects, thereby leading to superior performance.

Additionally, we compare our method with other powerful image restoration backbones, including Uformer~\cite{wang2022uformer}, Restormer~\cite{zamir2022restormer}, WeatherDiffusion~\cite{ozdenizci2023restoring} and DiT~\cite{Peebles2023DiT}. The experimental results are shown in Table~\ref{tab_quan}. It can be observed that existing image restoration methods trained using synthetic datasets struggle to restore haze images. The main reason is that their learning strategies, based on synthetic datasets, struggle to develop strong priors. This results in inferior performance.

\subsection{Qualitative Evaluation on RealNightHaze} 

Figure \ref{fig_exp1} presents a qualitative comparison on RealNightHaze. 
It is evident that existing nighttime image dehazing methods cannot effectively recover a clean background in challenging areas, such as those with low-light and glow effects. The main reason is that there is a large domain gap between their synthetic datasets and nighttime haze degradation. 
In contrast, our NightHaze method not only removes nighttime haze effects but also enhances the visibility of images. This is because our self-prior learning effectively forces our network to learn strong priors. 
Similar to MAE, our self-prior learning is also built on augmentation. In our case, we utilize two key challenging aspects of nighttime images as augmentation factors: light effects and noise. Following the same principle as MAE, we define and employ severe augmentation to obtain a strong prior, which significantly enhances performance.  
Furthermore, we employ a self-refinement module to further refine our model through learning from real haze images. To be more specific, our teacher model generates high-confidence predictions from real haze images to guide the student model's learning. This method enables our model to gradually adapt to real-world high-confidence predictions, thereby addressing more challenging scenarios.

\begin{figure}[t]
    \centering
    \subfloat[Inputs]
    {\includegraphics[width=2.5cm, height=2.8cm]{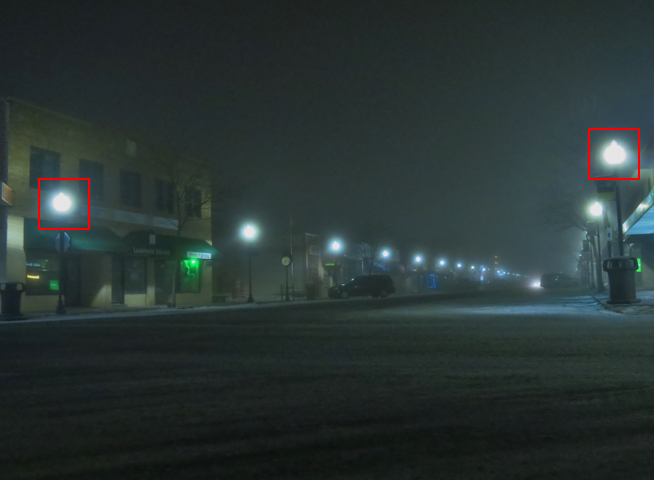}}\hspace{0.5pt}
    \subfloat[PL]
    {\includegraphics[width=2.5cm, height=2.8cm]{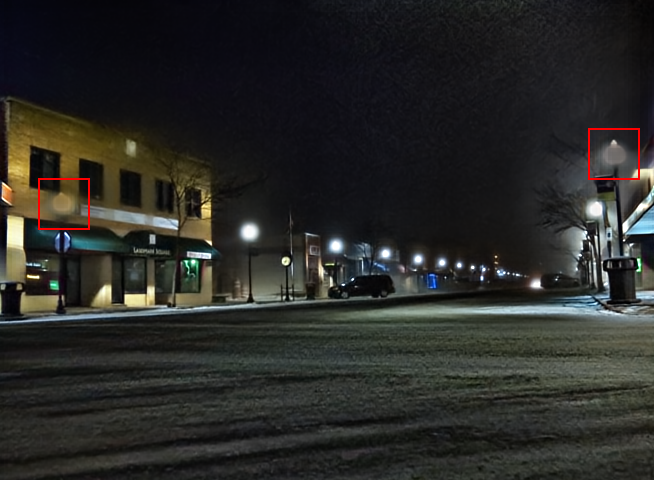}}\hspace{0.5pt}
    \subfloat[PL+SR]
    {\includegraphics[width=2.5cm, height=2.8cm]{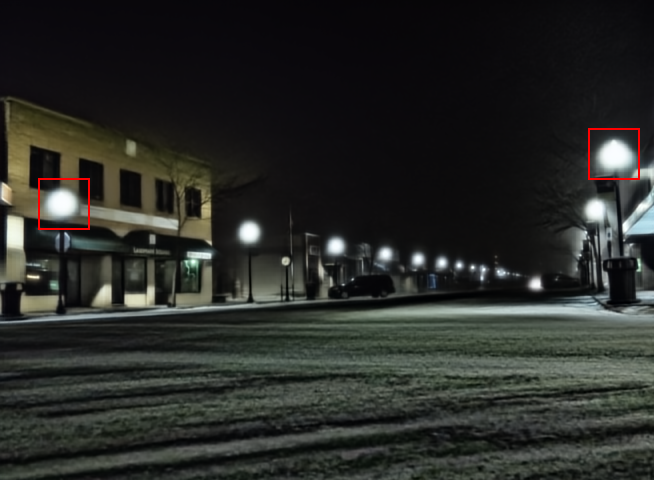}}\hspace{0.5pt}
    
    
    \caption{Ablation studies on real-world haze images. `PL' and `SR' are self-prior learning and self-refinement, respectively. (a) Input  (b) Results obtained using our self-prior learning. (c) Results obtained using our self-prior learning and self-refinement. Zoom in for better visualisation.
    }
    \label{fig_qual_Ablation}
\end{figure}

\subsection{Ablation Studies}

To verify the effectiveness of our key ideas, we conduct ablation studies on the RealNightHaze dataset. The experimental results are shown in Figure~\ref{fig_qual_Ablation} and Table~\ref{tab_abl}. 

\noindent{\textbf{Analysis of Self-Prior Learning}} 
To verify the effectiveness of our self-prior learning, we use different ratios to mix severe and non-severe augmentation. The experimental results are shown in Table~\ref{tab_abl}. It can be observed that our backbone with non-severe augmentation cannot sufficiently enhance the visibility of haze images.  
The MUSIQ and TRES scores are 44.67 and 62.18, outperforming the original haze images' scores by 2.55 and 4.24. 
In contrast, our backbone with severe augmentation achieves significant performance improvement. The main reason for this is that severe augmentation yields strong priors that are robust to real haze.

\noindent{\textbf{Analysis of Self-Refinement}} We conduct ablation studies to verify the effectiveness of our self-refinement. The experimental results are shown in Figure~\ref{fig_qual_Ablation}. Although our self-prior learning can effectively enhance the visibility of night haze images, it still leaves some artifacts caused by over-suppression. For example, the light sources in Figure~\ref{fig_qual_Ablation}(b) are over-suppressed. Figure~\ref{fig_qual_Ablation}(c) demonstrates that our self-refinement further refines our outputs.

\section{Conclusion}
This paper brings the MAE-like framework to nighttime image restoration and shows that severe augmentation during training yields strong priors that are robust to real-world night haze effects. 
We present NightHaze, a novel nighttime dehaizng method with self-prior learning. 
%
Our self-prior learning is simple: we intentionally augment clear nighttime images by blending them with various light effects including glow and adding noise. Specifically, we blend clear images with various light maps containing artificial glow and add noise to these blended images. 
Inspired by MAE, which indicates that the severity of augmentation is key to obtaining strong priors, we define severe augmentation using our night-specific augmentations. We show that severe augmentation in training yields strong priors that are robust to real-world haze. 
Although our self-prior learning can effectively handle haze effects, it still leaves some artifacts caused by over-suppression.
Our self-refinement is proposed to address this issue. It is built on the semi-supervised teacher-student framework. Unlike traditional teacher-student frameworks, we incorporate non-reference metrics to monitor each update of our student model. This approach allows us to prevent inaccurate knowledge transfer from the student to the teacher.
Extensive experimental results demonstrate that our NightHaze achieves a significant performance improvement on real-world nighttime hazy datasets.

\bibliography{aaai25}

\end{document}